\documentclass{article}

\usepackage{natbib}
\usepackage[left=1.25in, top=1in, bottom=1in, right=1.25in]{geometry}
\usepackage[utf8]{inputenc}
\usepackage[T1]{fontenc}   

\usepackage{url}           
\usepackage{booktabs}     
\usepackage{amsfonts}     
\usepackage{nicefrac}       
\usepackage{microtype}      

\usepackage{amsmath}
\usepackage{amsthm}
\usepackage{graphicx}

\usepackage{subfig}
\usepackage[inline]{enumitem}
\usepackage{stmaryrd}
\usepackage[mathscr]{eucal}
\usepackage{amsmath}
\usepackage[export]{adjustbox}

\usepackage{colortbl}

\usepackage[dvipsnames]{xcolor}

\newtheorem{definition}{Definition}

\newtheorem{theorem}{Theorem}
\newtheorem{lemma}[theorem]{Lemma}

\usepackage{algorithm}
\usepackage{algorithmic}

\newcommand{\ie}{\emph{i.e.}}

\usepackage{tikz}
\usetikzlibrary{decorations.pathreplacing,calc}
\newcommand{\tikzmark}[1]{\tikz[overlay,remember picture] \node (#1) {};}

\newcommand*{\AddNote}[4]{%
    \begin{tikzpicture}[overlay, remember picture]
        \draw [decoration={brace,amplitude=0.5em},decorate, thick]
            ($(#3)!(#1.north)!($(#3)-(0,1)$)$) --  
            ($(#3)!(#2.south)!($(#3)-(0,1)$)$)
                node [text width=2.5cm, pos=0.5, anchor=west] {#4};
    \end{tikzpicture}
}%

\DeclareMathOperator{\Var}{Var}
\DeclareMathOperator{\Cov}{Cov}

\usepackage[framemethod=tikz]{mdframed}
\newmdenv[
  linecolor=cyan,
  linewidth=2pt,
  roundcorner=5pt,
  innertopmargin=0pt,
  innerbottommargin=0pt,
]{myframe}

\title{Scaling Hierarchical Agglomerative Clustering to Billion-sized Datasets}

\author{%
  Baris Sumengen \\ 
  Google Research\\
  \texttt{sumengen@google.com} \\
  \and
  Anand Rajagopalan \\ 
  Google Research\\
  \texttt{anandbr@google.com} \\
  \and
  Gui Citovsky \\
  Google Research\\
  \texttt{gcitovsky@google.com} \\
  \and
  David Simcha \\
  Google Research\\
  \texttt{dsimcha@google.com} \\
  \and
  Olivier Bachem \\
  Google Research\\
  \texttt{bachem@google.com} \\
  \and
  Pradipta Mitra \\
  Google Research\\
  \texttt{ppmitra@google.com} \\
  \and
  Sam Blasiak \\
  Google Research\\
  \texttt{samblasiak@google.com} \\
  \and
  Mason Liang \\
  0x Labs\\
  \texttt{mason@0xproject.com} \\
  \and
  Sanjiv Kumar \\
  Google Research\\
  \texttt{sanjivk@google.com} \\
}
\date{}
\begin{document}

\maketitle

\begin{abstract}
Hierarchical Agglomerative Clustering (HAC) is one of the oldest but still most widely used clustering methods. However, HAC is notoriously hard to scale to large data sets as the underlying complexity is at least quadratic in the number of data points and many algorithms to solve HAC are inherently sequential. In this paper, we propose {Reciprocal Agglomerative Clustering (RAC)}, a distributed algorithm for HAC, that uses a novel strategy to efficiently merge clusters in parallel. We prove theoretically that RAC recovers the exact solution of HAC. Furthermore, under clusterability and balancedness assumption we show provable speedups in total runtime due to the parallelism. We also show that these speedups are achievable for certain probabilistic data models. In extensive experiments, we show that this parallelism is achieved on real world data sets and that the proposed RAC algorithm can recover the HAC hierarchy on billions of data points connected by trillions of edges in less than an hour.
\end{abstract}

\section{Introduction}
The recent, unprecedented rise in large data sets has been largely fueled by the growth in unstructured and unlabeled data. A fundamental method for attaching structure to these types of datasets is through clustering, and advancing the state of the art in clustering continues to attract both practical and theoretical interest. Despite many advances, scaling clustering methods to massive data sets remains a hard task. 

Hierarchical Agglomerative Clustering (HAC) is one of the oldest but still most widely used clustering methods. Starting from a state in which every datapoint is in its own cluster, it successively merges pairs of clusters together to produce a hierarchy. The popularity of HAC stems from its ease of implementation coupled with a variety of useful properties. Unlike popular parametric methods such as $k$-Means, HAC can generate flat clusterings by halting the merging process at a desired point, rather than requiring the number of clusters to be specified in advance. HAC's non-parametric nature also frees the user from the need to make assumptions about data distributions. Parametric clustering approaches can fail in high dimensional spaces since the underlying geometric shapes, \ie, Voronoi partitions in the case of $k$-Means, become meaningless.
HAC is also easily tunable through the use of different ``linkage functions'' that control how dissimilarities are updated when clusters merge. Finally, HAC enjoys the reputation for producing quality clusterings, and this behavior recently gained theoretical support~\citep{moseley2017approximation}.

Scaling HAC has proven to be challenging. The sequentiality and random memory access requirements of HAC make it hard to parallelize computation, a critical aspect where data sets do not fit on a single machine. Previous efforts have been limited to inputs containing at most hundreds of thousands of datapoints \citep{jeon2015multi}. Single linkage HAC is the exception because of its unique connection to the minimum spanning tree problem \citep{rammal1985degree}, but has the drawback that it produces hierarchies which are highly sensitive to outliers \citep{williams1966multivariate}. Another scalable approach to hierarchical clustering is affinity clustering \citep{bateni2017affinity}, which is capable of scaling to billions of datapoints, but produces hierarchies which differ from HAC.

We start with a brief introduction to HAC in Section \ref{sec:hac}. In Section \ref{sec:rac:parallel} we describe a massively scalable HAC algorithm (Algorithm \ref{alg:rac-high-level}), which we call Reciprocal Agglomerative Clustering (RAC). RAC computes the exact HAC hierarchy for any reducible linkage (defined in Section \ref{sec:hac}, examples include average and complete linkage), and is scalable to billions of datapoints connected by trillions of edges. In Section \ref{sec:proofs}, we give a rigorous proof of its correctness in Theorem \ref{thm:rac_correctness}. We note that while idea of merging reciprocal nearest neighbors appears in \cite{murtagh1983survey}, there is no proof of correctness, speedup theory, or any details on parallel implementation. We next give a negative example (Theorem \ref{thrm:nncomputation}) where RAC's parallelism does not guarantee speedup. We complement this result however by giving theoretical guarantees of runtime speedups in two scenarios: (i) Assuming the datapoints satisfy a notion of \emph{stability} as defined in \citet{awasthi2014local}, and (ii) An average-case analysis of speedup when the graph of the data is an instance of a natural class of random graph models. We provide details of our implementation in Section \ref{sec:implementation}, and in Section \ref{sec:experiments} we use RAC to cluster billion-sized real world data sets and give scaling results with respect to CPUs and machines.

\section{Hierarchical Agglomerative Clustering}
\label{sec:hac}

Hierarchical agglomerative clustering (HAC) produces a hierarchy of clusters, providing relationships between groups of datapoints at increasingly granular levels. An important parameter of this algorithm is the linkage function, which we will denote as $W$, which defines the dissimilarity between two clusters of datapoints as a function of the dissimilarities between their constituent elements. Many commonly used clustering algorithms, e.g. SLINK~\citep{sibson1973slink}, CLINK~\citep{defays1977efficient}, and UPGMA~\citep{sokal1958statistical} correspond to HAC with a specific linkage function. The three previously mentioned algorithms use single, complete, and average linkages, which are defined in Table~\ref{tbl:linkage_functions}.

\begin{table}[t]
  \caption{Linkage functions}
\label{tbl:linkage_functions}
  \centering
  {
      \begin{tabular}{lll}
      \toprule
      				& Definition 								& Update formula 		 	\\
                    \midrule
      	Single 		& $\min_{a\in A, b\in B} W_{a, b}$  		& $\min(W(A,C), W(B,C))$					\\
      	Complete 	& $\max_{a\in A, b\in B} W_{a, b}$ 			& $\max(W(A,C), W(B,C))$					\\
      	Average 	& $\sum_{a\in A, b\in B}W_{a, b} / |A||B|$ 	& $\frac{|A|W(A,C) +|B|W(B,C)}{|A| + |B|}$	\\
        \bottomrule
      \end{tabular}
  }
\end{table}

Starting from a state in which every data point is in a singleton cluster, HAC creates a hierarchy by sequentially combining the two clusters which are most similar according to the linkage function. This process is repeated until all of the original data points have been coalesced into a single cluster, and the unordered list of mergers, representing the cluster hierarchy, is returned (Algorithm~\ref{alg:hac}).

\begin{algorithm}[t]
\caption{Hierarchical Agglomerative Clustering (HAC)}
\label{alg:hac}
\begin{algorithmic}[1]
\REQUIRE{Set of clusters, $\mathcal{C}$, and a set of merges, $\mathcal{M}$, with default value $\emptyset$}
\IF{$|\mathcal{C}| = 1$}
  \STATE {\bfseries Return} $\mathcal{M}$
\ENDIF
  \STATE $A, B \gets$ Pair of distinct clusters in $C$ which minimize $W(A,B)$
  \STATE $\mathcal{C} \gets \{A \cup B\} \cup \mathcal{C}\setminus \{A, B\}$
  \STATE $\mathcal{M} \gets \mathcal{M} \cup \{\{A,B\}\}$
  \STATE {\bfseries Return} $\text{HAC}(\mathcal{C}, \mathcal{M})$
\end{algorithmic}
\end{algorithm}
The linkage functions defined in Table \ref{tbl:linkage_functions} all belong to the more specific class of reducible, \emph{Lance-Williams}~\citep{lance1967general} linkage functions. These linkage functions have the property that the dissimilarities between clusters may be computed recursively in terms of dissimilarities between their refinements, i.e. given clusters $A, B$, and $C$, along with all their pairwise dissimilarities, we can compute $W(A\cup B, C)$ in $O(1)$ time rather than by iterating over all of individual pairs of datapoints in the two sets, which would require $O(|C|(|A|+|B|))$ time.

\emph{Reducible} linkage functions have the property that merging two clusters does not make the resulting union more similar to any other cluster, i.e. for any three disjoint clusters $A, B,$ and $C$, $W(A\cup B, C)\geq \min(W(A,C), W(B,C))$. Centroid linkage, where the dissimilarity between clusters is the distance between their centroids, is an example of a linkage function which is not reducible. If the linkage function used in HAC is reducible, then the dissimilarities between the clusters which are being merged in Algorithm~\ref{alg:hac} will be non-decreasing.

There has been work on parallelizing portions of the HAC algorithm in the past. \citet{cathey2007exploiting} keep the merges sequential but parallelizes the work done in each merge, e.g. finding the least dissimilar clusters and updating the cluster dissimilarities. \citet{jeon2015multi} parallelizes the nearest neighbor chain algorithm \citep{murtagh1984complexities} by using multiple threads to follow different chains but collisions between these chains could potentially create thread contention. In both cases, the algorithms were demonstrated on single machines for datasets with under a million datapoints.

\section{Reciprocal Agglomerative Clustering}

\label{sec:rac:parallel}

When the HAC procedure is run using a reducible linkage function, the order in which clusters are merged can be relaxed \citep{de1980classification,murtagh1983survey,rasmussen1989efficiency}.
In particular, rather than merging the pair of clusters whose dissimilarity globally minimizes $W(A,B)$, with a reducible linkage function, it is possible to merge any pair of clusters, $A$ and $B$, that are \emph{reciprocal nearest neighbors}. Reciprocal nearest neighbors are clusters that are nearest neighbors of one another, i.e., there exists no other cluster $C\in\mathcal{C}$ such that $W(A,C) < W(A,B)$ or $W(B,C) < W(A,B)$.
The key idea behind the \emph{nearest neighbor chain algorithm} is to follow the nearest neighbor graph to identify such reciprocal nearest neighbors and to \emph{sequentially} merge them.

Inspired by this idea of reciprocal nearest neighbors, we propose a distributed algorithm called \emph{Reciprocal Agglomerative Clustering} (RAC) to scale HAC to billions of data points. We describe a high level overview of RAC in this Section. RAC proceed in rounds where in each round, we merge \emph{all} reciprocal nearest clusters \emph{in parallel} on a set of machines instead of sequentially following a nearest neighbor chain.

Algorithm \ref{alg:rac-high-level} presents a high level overview of RAC. Each round of RAC proceeds in two phases of \emph{finding reciprocal nearest neighbors} and \emph{merging them}, each of which is parallelized. The merge phase may be further subdivided into updating cluster dissimilarities and updating nearest neighbors (see Section~\ref{sec:implementation} for a full distributed implementation).

1. \textbf{Find Reciprocal Nearest Neighbors:}
Given a set of clusters $\mathcal{C}$, we find and return in parallel all merges: pairs of clusters $\{C_i, D_i\}$ such that $C_i$'s nearest neighbor is $D_i$ and vice-versa.\\
2a. \textbf{Update Cluster Dissimilarities}:
Once we merge the pairs of reciprocal nearest neighbor clusters, we update in parallel the dissimilarities of all pairs of clusters that have been changed and delete the original clusters that were merged. \\
2b. \textbf{Update Nearest Neighbors}:
For each cluster $C$ that belongs to a reciprocal nearest neighbor pair, and for each of $C$'s neighbors $N$, we find and update in parallel the new nearest neighbor of $N$ (since after merging $C$ with its reciprocal nearest neighbor, $C$ may no longer be $N$'s nearest neighbor).

\begin{algorithm}
\caption{Reciprocal Agglomerative Clustering (RAC)}
\begin{algorithmic}[1]
\STATE \textbf{Require: }Set of clusters, $\mathcal{C}$, and a set of merges, $\mathcal{M}$, with default value $\emptyset$
\STATE $\mathcal{M} \gets \mathcal{M} \cup \text{Find Reciprocal Nearest Neighbors}(C)$
\STATE $\text{Update Cluster Dissimilarities}(C)$ \tikzmark{right} \tikzmark{top}
\STATE $\text{Update Nearest Neighbors}(C)$ \tikzmark{bottom}
\STATE {\bfseries Return} RAC($\mathcal{C}, \mathcal{M}$)
\AddNote{top}{bottom}{right}{\; Merge}
\label{alg:rac-high-level}
\end{algorithmic}
\end{algorithm}

\vspace{-10pt}
\section{Theoretical Results}
\label{sec:proofs}

In this section we first prove the correctness of RAC. RAC, as we have seen, proceeds in rounds, where each round performs a set of parallel merges. RAC can hope to perform significantly better than sequential implementations of HAC only if the number of rounds is significantly smaller than $n$ (the number of input datapoints). We first present a negative example showing that in the worse case, RAC can take $O(n)$ rounds even when the cluster tree is of depth $O(\log n)$ (intuitively, a case with high parallelism). We then complement this by giving theoretical guarantees in two scenarios where the number of rounds is $O(\log n)$: We provide a characterization where the number of rounds of RAC is the same as the tree height, providing an explanation of its excellent performance in practical scenarios. We also show that RAC can achieve a provable average-case speedup in a probabilistic setting when the edges of the data are drawn from a random graph model. Lastly we provide a connection between the number of merges per-round and the overall complexity of RAC, proving that RAC can perform in near linear time under reasonable assumptions (in the experimental section we provide empirical proof that these assumptions are reasonable for many real applications). 

\subsection{Proof of Correctness}
We start with the correctness proof,
showing that given a reducible linkage and a set of clusters $\mathcal{C}$,
$\textsc{HAC}(\mathcal{C}) = \textsc{RAC}(\mathcal{C})$.
The main idea in this proof is that at any state of HAC, there exists a set of pairs of clusters that are reciprocal nearest neighbors.
We will first show that given a pair of clusters that are reciprocal nearest neighbors, HAC will merge them.
We then show that the \textbf{order} in which HAC merges these reciprocal nearest neighbors does not matter.
The combination of these results implies that all reciprocal nearest neighbors may be merged in parallel,
with the resultant clusters matching the ones produced by HAC.
\begin{theorem}
\label{thm:rac_correctness}
For reducible linkages $W$, Hierarchical Agglomerative Clustering and Reciprocal Agglomerative Clustering produce the same result. That is, for a set of clusters $\mathcal{C}$, $\textsc{HAC}(\mathcal{C}) = \textsc{RAC}(\mathcal{C}).$
\end{theorem}
\begin{proof}
We prove this by strong induction on $|\mathcal{C}|$. The claim holds for $|\mathcal{C}|=1$. We will use the following two lemmas:
\begin{lemma}
\label{rac_equiv:rac_merge_happens_in_hac}
Let clusters $A, B\in \mathcal{C}$ be a pair of reciprocal nearest neighbors and let $\epsilon = W(A, B)$. Then the merge between $A$ and $B$, $[A, B]$, will appear in $\textsc{HAC}(\mathcal{C})$.
\end{lemma}
\begin{proof}
Let $\mathcal{C}^*$ be any non-empty subset of $\mathcal{C}$ which does not contain $A$ or $B$ and let $D$ be the union of all the clusters in $\mathcal{C}^*$. Using the fact that $W$ is reducible and that $\epsilon < W(A,X)$ for any $X\in \mathcal{C}$ that is not $A$ or $B$, we have that $W(A,D) = W\left(A,\cup_{X\in \mathcal{C}^*}X\right) 
    \geq \min_{X\in \mathcal{C}^*}\left(W(A, X)\right)
    > \epsilon.$ 
This strict inequality means that $A$ cannot merge with $D$ while $B$ is still unmerged. A similar argument applies to $B$. Since all clusters in $\mathcal{C}$ will eventually be merged, the only possibility remaining is that $A$ and $B$ will be merged.
\end{proof}

We now show that the order in which HAC merges reciprocal nearest neighbors in $\mathcal{C}$ does not matter.

\begin{lemma}
\label{rac_equiv:same_later_merges}
Let $A, B\in \mathcal{C}$ be two clusters which are merged at some point in $\text{HAC}(\mathcal{C})$. Let $\mathcal{C'} = \{A \cup B\} \cup \mathcal{C} \setminus \{A, B\}$. Then $\textsc{HAC}(\mathcal{C}, \mathcal{M}) = \textsc{HAC}(\mathcal{C'}, \mathcal{M}\cup [A, B]).$
\end{lemma}

\begin{figure}[h!]
  \begin{center}
  \includegraphics[scale=0.6]{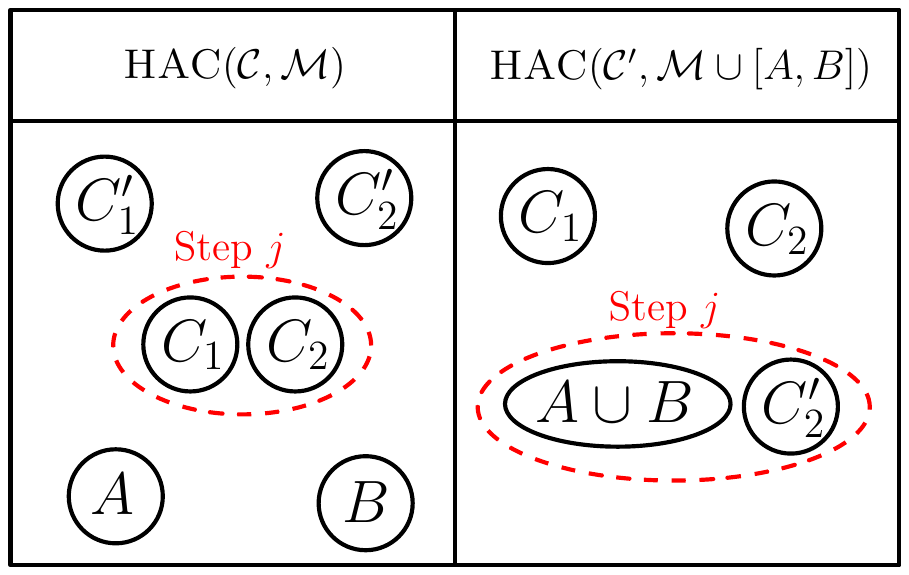}
  \caption{Step $j$ in $\textsc{HAC}(\mathcal{C}, \mathcal{M})$ and in $\textsc{HAC}(\mathcal{C'}, \mathcal{M}\cup [A, B])$}
  \label{fig:same_later_merges}
  \end{center}
\end{figure}
\vspace{-15pt}
\begin{proof}
Let $\mathcal{C}_k$ (resp. $\mathcal{C}'_k$) be the set of clusters produced after $k$ iterations of $\text{HAC}(\mathcal{C})$ (resp. $\textsc{HAC}(\mathcal{C}', \mathcal{M}\cup [A, B])$).
Let $s$ be the iteration in which the merge between $A$ and $B$ occurs in $\text{HAC}(\mathcal{C})$.
We will now show that $\mathcal{C}_s = \mathcal{C}'_s$.

Suppose by contradiction that $\mathcal{C}_s \neq \mathcal{C}'_s$.
Consider the first iteration $j < s$ in the respective procedures, s.t. $\mathcal{C}_j \neq \mathcal{C}'_j$. Let ($C_{1}$, $C_{2}$) (resp. ($C'_{1}$, $C'_{2}$)) be the clusters that were merged in the creation of $\mathcal{C}_j$  (resp. $\mathcal{C}'_j$), where $C_{1} \cup C_{2} \neq C'_{1} \cup C'_{2}$.
Since $\mathcal{C}_{j-1} = \mathcal{C}'_{j - 1} \setminus \{A \cup B\} \cup \{A, B\} $,
it must be the case that $C'_{1} = A \cup B$ or $C'_{2} = A \cup B$ (see Figure~\ref{fig:same_later_merges}).

Suppose without loss of generality that $C'_{1} = A \cup B$.
We know that $W(A \cup B, C'_{2}) < W(C_{1}, C_{2})$
since otherwise $C_{1}$ and $C_{2}$ would have merged in the creation $\mathcal{C}'_j$ (right side of Figure~\ref{fig:same_later_merges}).
Further, we know that $W(C_{1}, C_{2}) < W(A, C'_{2})$
since otherwise $A$ and $C'_{2}$ would have merged in the creation of $\mathcal{C}_j$ (left side of Figure~\ref{fig:same_later_merges}).
By similar logic, we have that $W(A \cup B, C'_{2}) < W(B, C'_{2})$.
Thus, $W(A \cup B, C'_{2}) < \min(W(A, C'_{2}), W(B, C'_{2}))$,
a contradiction to $W$ being reducible.

\end{proof}

To finish the proof of Theorem \ref{thm:rac_correctness}, assume $|\mathcal{C}|\geq 2$. Let $\mathcal{R} = \{R_1, R_2, R_3, R_4, \dots \}$, be a set of clusters in $\mathcal{C}$ where $\{R_{2i - 1}, R_{2i}\}, i = 1, 2, 3, \dots$ are reciprocal nearest neighbor pairs. Let $\mathcal{R'} = \{R_1 \cup R_2, R_3 \cup R_4, \dots\}$ be their unions. By repeatedly applying the results from the above two lemmas, the inductive hypothesis and finally, using the definition of RAC, we get that
\begin{align*}
	\text{HAC}(\mathcal{C}) 
    & = \text{HAC}(\mathcal{R'}\cup \mathcal{C} \setminus \mathcal{R}, \{[R_1, R_2], [R_3, R_4] \dots\}) \\&
	= \text{RAC}(\mathcal{R'}\cup \mathcal{C} \setminus \mathcal{R}, \{[R_1, R_2], [R_3, R_4] \dots\})  \\&
	=  \text{RAC}(\mathcal{C})
\end{align*}
\end{proof}

\subsection{Speedup guarantees} We start by analyzing the number of merge rounds required by RAC.
First, a negative result.
Clearly, the number of rounds RAC requires is bounded from below by the height of the resulting dendrogram. One might hope that this bound is tight, in which we can use the tree height to characterize the parallelism inherent in a clustering problem. However, we show below there are inputs which require $O(n)$ rounds of RAC, even though the resulting dendrogram has only $O(\log(n))$ height. 

\begin{theorem}
\label{thrm:nncomputation}
For any $n\in \mathbb{N}$, set $X =\{1 + \epsilon, 2 + 4\epsilon, 3 + 9\epsilon, \dots, 2^n + 2^{2n} \epsilon \}$, where $\epsilon = 2^{-4n}$.
Then $\textsc{RAC}(X)$ returns a dendrogram of height $n$, but requires $\Omega(2^n)$ rounds of reciprocal nearest neighbor merges to complete.
\emph{[Proof in supplementary materials.]}
\end{theorem}

\subsubsection{Stable Trees}
Notwithstanding this negative result, we can show that if the average linkage cluster tree has certain "clusterability" properties, then the number of rounds RAC requires is the same as the tree height. In \cite{awasthi2014local}, the authors introduce clusterability notion for flat clusters called \emph{stability}. We adapt this definition for cluster trees as follows:
\begin{definition}
A given cluster tree $T$ is \emph{stable} if for any non-overlapping nodes $X, Y \in T$ (i.e. the set of leaves of $X$ and $Y$ have empty intersection), and any $A \subset X$, $B \subseteq Y$ $d(A, X \setminus A) < d(A, B)$

\label{perf:def:stability}
\end{definition}

\begin{theorem}
\label{theorem:RAC_rounds}
 On stable cluster trees, RAC completes in a number of rounds equal to the tree height.
\end{theorem}

\begin{proof}
Let the set of datapoints in the cluster tree be $P$, and $h$ be the height of the tree. Define 
$\rho(X)$ to be the parent of $X$ in the cluster tree.
Define a \emph{Leveled Clustering} at level l,  $C_l$ for $0 \leq l < h$ of the tree recursively as follows: 
for a level $l$, let $M_l = \{X \in C_l : \exists Y \in C_l, \rho(X) = \rho(Y)\}$, i.e., nodes in $C_l$ that merge with another node in $C_l$, and let 
$$     C_l = 
\begin{cases}
    P,& \text{if } l = 0\\
    C_{l-1} \setminus M_{l-1} \cup \{\rho(X): X \in M_{l-1}\},              & \text{otherwise}
\end{cases}
$$

We claim that the state of the RAC algorithm after $l$ rounds of merging is $C_l$. Note that this implies that the number
of rounds required is $h$, proving the theorem. Let us prove this recursively, the base case being obvious. 
For the inductive step, assume the cluster state is $C_l$. Let $|M_l| = 2 \cdot M$ for an integer $M$ (it is clear that  $|M_l|$ is even) and let the members of $M_l$ be  $m_0, m_1 \ldots m_{2M-1}$. Assume without loss of generality that $m_{2i}$ and $m_{2\cdot (i+1)}$ are siblings (for all $i \leq M - 1$), i.e., $\rho(m_{2i}) = \rho(m_{2(i+1)})$. Since we
know that RAC produces the correct cluster tree, all we need to demonstrate is that all $m_{2i}$ and $m_{2\cdot (i+1)}$ get merged simultaneously in the next step of RAC, or equivalently, that $m_{2i}$ and $m_{2\cdot (i+1)}$ are reciprocal nearest neighbors. Take any other node $Y \in C_l$. Set $X = m_{2i} \cup m_{2\cdot (i+1)}$, $A = m_{2i}$ and $B = Y$, and by the
definition of stability, we see that $d(m_{2i}, m_{2i+1}) < d(m_{2i}, Y)$. The same argument can be shown to imply that
$d(m_{2i}, m_{2i+1}) < d(m_{2i+1}, Y)$ and thus $m_{2i}$ and $m_{2\cdot (i+1)}$ are reciprocal nearest neighbors.
\end{proof}

\subsubsection{Probabilistic setting}
\label{subsec:prob}
We can extend the conclusion of Theorem \ref{theorem:RAC_rounds} to probabilistic settings which do not require the assumption of stability for the case of single linkage. In particular, we present a probabilistic graphical model in which the \emph{expected} number of merges in each round is a constant fraction of the number of clusters in that round. Moreover, we show that this implies that RAC terminates in $O(\log n)$ rounds both in expectation and with high probability. We start with a proof of the latter result. 

\begin{theorem}
\label{thm:logn-rounds}
Fix integer $n > 0$ and real $\alpha > 0$. Let $(X_k)_{k\geq 0}$ be a sequence of random variables with $X_0 = n$ and $X_{k+1} = 1$ if $X_k = 1$ and $X_{k+1} = X_k - Z_k$ otherwise, with $Z_k$ satisfying the following conditions:
\begin{enumerate}[label=(\roman*)]
\vspace{-10pt}
\item $Z_k$ is integer valued and $1\leq Z_k\leq X_k-1$ and
\vspace{-5pt}
\item $E[Z_k | X_k] \geq \alpha X_k$.
\end{enumerate}
\vspace{-10pt}
Let $\tau = \arg\min_{k}\{X_k=1\}$. Then $\tau = O(\log n)$ in expectation and with high probability.
\end{theorem}
The application of Theorem \ref{thm:logn-rounds} to RAC is as follows: Let $X_k$ be the number of clusters and $Z_k$ be the number of merges in round $k$ respectively. Then, if we can show that at least $\alpha$ fraction of merges occur in each round in expectation, then RAC terminates in $O(\log n)$ rounds of merges.
\begin{proof}
Begin by letting $Y_k = X_k/(1-\alpha)^k$. We use the optional stopping theorem \cite{doob1953stochastic} applied to the $Y_k$'s to prove the result. In this context, the theorem states that if (a) $\tau$ is bounded and (b) $E[Y_{k+1}|Y_k] \leq Y_k$,
then $E[Y_{\tau}] \leq E[Y_0] = n$. Assuming for now that (a) and (b) hold, by Jensen's inequality, $E[Y_{\tau}] = E[1/(1-\alpha)^{\tau}] \geq 1/(1-\alpha)^{E[\tau]}$ which gives $E[\tau] \leq \log n/\log(1/(1-\alpha))$. The high probability guarantee follows from Markov's inequality:
\begin{align*}
    P[\tau > (1+\beta)\frac{\log n}{\log(1/(1-\alpha))}] &= P[1/(1-\alpha)^{\tau} > n^{1+\beta}] \\
    &\leq E[1/(1-\alpha)^{\tau}]/n^{1 + \beta}\\
    &\leq n^{1-\beta}.
\end{align*}
We turn to proving (a) and (b). To see (a), note that by
condition (i), $X_k$ decreases by positive integer values in each round, and hence $\tau \leq n-1$. (b) follows from the following calculation.
\begin{align*}
    E[Y_{k+1}|Y_k] &= E[Y_{k+1}|X_k]\\
    &= (1/(1-\alpha)^{k+1})E[X_{k+1}|X_k]\\
    &= (1/(1-\alpha)^{k+1})E[X_k-Z_k|X_k]\\
    &\leq (1/(1-\alpha)^{k+1})X_k(1-\alpha)\\
    &= Y_k.
\end{align*}

\end{proof}

We now apply this result to two probabilistic settings.\\
\textbf{Single Linkage, $1$-dimensional grid:}
Consider points $\{x_1, x_2, \ldots, x_n\}$ on the real axis with $x_1 < x_2 < \ldots < x_n$ such that the dissimilarities between consecutive points are sorted uniformly at random. Such a model can be generated by choosing the $x_i$'s iid on the unit interval $[0, 1]$ and then relabeling them in increasing order. Suppose after some number of rounds, $k$ clusters $C_1, C_2, \ldots, C_k$ remain with $k > 2$. It is clear that each $C_i$ consists of a subset of points with contiguous indices, thus $C_i = \{x_{j_i}, x_{j_i + 1}, \ldots, x_{j_i+ n_i}\}$. Denoting the dissimilarity between $C_i$ and $C_{i+1}$ by $d(C_i, C_{i+1})$, the probability that $C_i$ merges with $C_{i+1}$ is then given by $P[d(C_i, C_{i+1}) < d(C_{i-1}, C_i)]$ if $i = k-1$, $P[d(C_i, C_{i+1}) < d(C_{i+1}, C_{i+2})]$ if $i = 2$, and 
$P[d(C_i, C_{i+1}) < \min(d(C_{i-1}, C_i), d(C_{i+1}, C_{i+2}))]$ otherwise. Under the assumption of randomly sorted edge weights, this simplifies to 
\begin{equation*}   
P[C_i\text{ merges with }C_{i+1}] = 
     \begin{cases}
       1/3 &\quad\text{if }$i = 2, 3, \ldots, k-2$ \\
       1/2 &\quad\text{if }$i = 1$\text{ or }$i = k-1$.
     \end{cases}
\end{equation*}
By linearity of expectation, the number of merges is at least $k/3$ (if $k = 2$, then a single merge occurs which also satisfies this). Thus, we can apply Theorem \ref{thm:logn-rounds} with $\alpha = 1/3$.

\textbf{Single Linkage, Bounded Degree Probabilistic Graph:}
We can generalize the previous result as follows. Let $G = (V, E)$ be a weighted graph on $n$ vertices with the weights sorted at random and each vertex initially being a singleton cluster. Let $E(C_i, C_j)$ denote the edges between clusters $C_i$ and $C_j$. In a given round, the pair of clusters $C_i$ and $C_j$ merge if 
\begin{equation*}
\min_{e\in E(C_i, C_j)} e < \min(\min_{l\neq j}\min_{e\in E(C_i, C_l)}e, \min_{l\neq i}\min_{e\in E(C_l, C_j)}e),    
\end{equation*}
i.e. according to the Single Linkage rule. Suppose that after some number of rounds, $k$ clusters remain. We can compute the expected number of merges as follows. Let $d_{ij} = |E(C_i, C_j)|$ and let $d_i = \sum_{j\neq i}d_{ij}$. Then the probability that $C_i$ and $C_j$ merge is given by
\begin{equation*}
P_{ij} = \frac{d_{ij}}{d_i + d_j + d_{ij}}.    
\end{equation*}
By linearity of expectation, the expected number of merges is given by 
$M(G) = \sum_{i=1}^k\sum_{j=i+1}^kP_{ij}$.
Now suppose that the cluster graph is of bounded degree, i.e., $d_i \leq d = O(1)$ (as mentioned before, this is a reasonable assumption and supported by experiments). Note that since eventually all clusters merge, $d_i \geq 1$ for all $i$. Thus, $|\{\{i, j\}|d_{ij} \geq 1\}| \geq k-1$. For each such $(i, j)$, $P_{ij} \geq 1/(d_i + d_j) \geq 1/(2d)$ from which we have $M(G) \geq (k-1)/(2d) \geq k/(4d)$ and Theorem \ref{thm:logn-rounds} applies with $\alpha = 1/(4d)$.

\subsection{Runtime analysis}
We now study the time complexity of RAC which depends on the number of rounds and the complexity of each round. In each round, the algorithm needs to: a) Find reciprocal nearest neighbors b) Perform the resultant merges c) Update nearest neighbors for every remaining cluster. Finding the reciprocal nearest neighbors for every cluster is $O(n)$ operation, totaling $O(n^2)$ over all rounds. Merging a pair costs $O(n)$, and there are $O(n)$ merges across all rounds, gaining another $O(n^2)$. Finally, recomputing nearest neighbors is a $O(\log n)$ operation if we use a min-heap. However, in practice, we simply iterate over a unsorted list in $O(n)$ (due to its improved cache-locality whose benefits outweighs the theoretical downsides). Nevertheless, this gives us $O(n^2)$ cost per-round for an overall cost of $O(n^3)$, which is rather large. However, this makes really worst-case assumptions about the number of merges and character of the data distribution that do not hold in practice. In particular, as we will see in practice in the experimental section, many merges happen simultaneously and the number of rounds are much smaller than $n$. We characterize this in the following theorem.

\begin{theorem}
Assume that a  constant $\alpha$ fraction of nodes are merged in each round. Then, RAC runs in expected time $O(n^2)$ (as opposed to $O(n^3))$.
\label{thm:nsquared_runtime}
\end{theorem}
\vspace{-5pt}
\begin{proof}\renewcommand{\qedsymbol}{}
Let the runtime be $f(n)$. Then we have the following:
\vspace{-10pt}
\begin{align*}
f(n) &
	\propto \overbrace{\alpha n \cdot n}^{\text{merges}} + \overbrace{\left(\left(1-\frac{\alpha}{2}\right) n\right)^2}^{\text{update nearest neighbors}} + f\left(\left(1 - \frac{\alpha}{2}\right)n\right) \\&
	\leq n^2 \left(\alpha + \left(1 - \frac{\alpha}{2}\right)^2\right) \sum_{i=0}^\infty \left(1 - \frac{\alpha}{2}\right) ^ i = O(n^2).
\end{align*}
\end{proof}

In view of the probabilistic models in the Subsection \ref{subsec:prob}, we have a probabilistic analog of Theorem \ref{thm:nsquared_runtime}:
\begin{theorem}
For the probabilistic graph model of Subsection \ref{subsec:prob} with an \emph{expected} constant fraction of nodes merging in each round, RAC runs in expected time $O(n^2)$.
\emph{[Proof in supplementary materials.]}
\label{thm:nsquared_runtime_prob}
\end{theorem}

Now we argue that under
reasonable assumptions, the runtime of the distributed algorithm can be nearly linear. Consider a problem with $n$ datapoints where every cluster has a degree bounded by $k$ (for large datasets this is a reasonable assumption -- See Section \ref{sec:experiments}). Consider a round in which $m$ merges occur.
Then there are 
$O(\min(n, m \cdot k))$ non-merging nodes that have their neighborhoods changed. 
Assume a fraction, $\beta$, of these have their nearest neighbor updated (equivalently, their nearest neighbor was merged). Each of these need $O(k)$ time to find the new nearest neighbor.
Then the run-time of RAC can be decomposed as in Table \ref{tbl:runtime_breakdown}.

\begin{table}[h!]
\caption{Breakdown of run-times into phases.}
  \centering
  {
      \begin{tabular}{l|l|l}
      \textbf{Algorithm step} & \begin{tabular}{@{}c@{}}\textbf{Main} \\\textbf{resource}\end{tabular}  & \textbf{Run-time}  \\ \hline
      Find reciprocal nearest neighbors &  Network  & $O(n)$\\
      Send neighborhoods for mergers &  Network  & $O(m \cdot k)$\\
      Merge &  Compute   & $O(m \cdot k)$\\
      Get info for non-merge updates &  Network   & $O(m \cdot k)$\\
      Non-merge updates &  Compute   & $O(m \cdot k)$\\
      Update nearest neighbors &  Compute   &
      \begin{tabular}{@{}l@{}}$O(\beta\cdot $ \\$ \min(n,mk^2))$\end{tabular}
      \end{tabular}
  }
 \label{tbl:runtime_breakdown}
\end{table}

It is clear that for efficient runtimes we want $m$ to be large and  $\beta$ to be small. Assume as before that $m = \alpha \cdot n$. Then the complexity of the $i^{\text{th}}$ round (following arguments similar to Theorem \ref{thm:nsquared_runtime}) is $O(\alpha \cdot n \cdot k + \alpha \beta n k^2)$. If $\beta = O(k^{-1})$ then the total run time will be $O(n \cdot k)$. This assumption on $\beta$ means that a merge pair can change at most a constant number of their neighbors nearest neighbor. We show experimental support for this assumption in the next section. Assuming a parallelism factor of $P$ (in the form of machines/CPU etc), we can state the following:
\begin{theorem}
If $\alpha = \Omega(1)$ (constant fraction of nodes merge in each round) and $\beta = \Theta(\frac{1}{k}$) (a merge pair changes at most a constant number of non-merging nodes nearest neighbors), then the runtime of RAC with $P$ parallelism factor is $O(\frac{n \cdot k}{P})$.
\label{thm:nearly_linear_time}
\end{theorem}
\vspace{-10pt}
For very large datasets $k$ is often bounded by a few hundred or thousands, and $P$ can scale up to many hundreds. Thus in practical situations, we can often hit nearly linear time performance with RAC.

\vspace{-5pt}
\section{Implementation}
\label{sec:implementation}
In this section, we provide a detailed implementation of RAC. In RAC, clusters are distributed across machines and have the following methods and attributes:
\textbf{id} - a unique numeric identifier for the cluster,
\textbf{will\_merge} - a boolean flag which indicates whether the cluster will be merging in the current round,
\textbf{nn} (abbreviation for \textbf{nearest\_neighbor}) - a cached reference to the closest cluster,
\textbf{neighbors} - the set of clusters with dissimilarities to the current cluster,
\textbf{update\_dissimilarity(other\_cluster, new\_dissimilarity)} - updates the stored dissimilarity value between this cluster and the other cluster to new value,
\textbf{get\_dissimilarity(other\_cluster) }- returns the cached value of the dissimilarity between this cluster and the other cluster, and 
\textbf{update\_nearest\_neighbor()} -  iterates over this cluster's neighbors to find and cache the reference to the neighboring cluster that this cluster is closest to.

In each round of RAC, we first find all reciprocal nearest neighbors and then perform the merges in parallel. These steps are repeated until no merges are possible. Between each step, we wait for all machines to finish.

We perform these parallel merges efficiently by avoiding contention as follows: for parallel merging clusters, $A\cup B$ and $C \cup D$, the value of $W(A \cup B, C \cup D)$ will be computed twice, once for $A\cup B$ and once for $C \cup D$. However, neither process needs to wait on the other. In practice, we find that this results in much greater throughput.

When two clusters are to be merged, the machine owning the cluster with the lower id will be responsible for the merge operation and will own the merge result. 
In practice, the cluster with the lower id will be overwritten by the merge result, and the cluster with the higher id will be deleted. 


Because the clusters are sharded over machines, information about a cluster's nearest neighbor, such as its nearest neighbor, can typically only be found via a remote call. Rather than blocking on this, we buffer remote calls and only issue them in batches. This allows us to pipeline the computation and communication processes, which in turn allows us to scale the number of machines. For simplicity, Algorithm \ref{alg:rac-high-level} does not show batching of remote calls. 

Algorithm \ref{alg:rac-high-level}, along with the following procedures used in the algorithm, provide a full implementation of RAC.

\begin{algorithm}[H]
\setcounter{algorithm}{0}
\caption*{\centering \textbf{Find Reciprocal Nearest Neighbors}}
\label{alg:find_rnns}
\begin{algorithmic}[1]
\STATE \textbf{Require: }Set of clusters, $\mathcal{C}$
\STATE $\mathcal{M} \gets \emptyset$
\COMMENT {Set of merges to return}
\IF{$|\mathcal{C}|=1$}
	\STATE {\bfseries Return} $\mathcal{M}$
\ENDIF
\WHILE{$C\in \mathcal{C}$} 
	\STATE $C.$will\_merge$\gets C$.nn.nn \textbf{is} $C$
\ENDWHILE 
\FOR{$C\in \mathcal{C}: C$.will\_merge and $C.\text{id} < C$.nn.id}
	\STATE $\mathcal{M} \gets \mathcal{M}\cup \{ \{C, C.\text{nn}\} \}$
\ENDFOR
\STATE {\bfseries Return} $\mathcal{M}$
\end{algorithmic}
\end{algorithm}

\vspace{-3mm}

\begin{algorithm}[H]
\caption*{\centering \textbf{Update Cluster Dissimilarities}}
\label{alg:update_cds}
\begin{algorithmic}[1]
\STATE \textbf{Require: }Set of clusters, $\mathcal{C}$
\WHILE{$C\in \mathcal{C}: C$.will\_merge \textbf{and} $C.\text{id} < C$.nn.id} 
	\STATE $\mathcal{N}\gets \emptyset$
	\FORALL{$C'\in C.\text{neighbors} \cup C.\text{nn}.\text{neighbors}$}
        \IF{$C'.\text{will\_merge}$ \textbf{ and } $C'\text{.id}< C'$.nn.id}
            \STATE $w\gets W(C\cup C.\text{nn}, C' \cup C'.\text{nn})$
            \STATE $C.\text{update\_dissimilarity}(C', w)$
            \STATE $\mathcal{N}\gets \mathcal{N} \cup \{C'\}$
        \ELSIF{\textbf{not }$C'.\text{will\_merge}$}
            \STATE $w\gets W(C\cup C.\text{nn}, C')$
            \STATE $C.\text{update\_dissimilarity}(C', w)$
            \STATE $\mathcal{N}\gets \mathcal{N} \cup \{C'\}$
        \ENDIF
	\ENDFOR
    \STATE $C.\text{neighbors}\gets \mathcal{N}$
\ENDWHILE 
\WHILE{$C\in \mathcal{C}: C$.will\_merge \textbf{and} $C.\text{id} > C$.nn.id} 
	\STATE Delete $C$
\ENDWHILE 
\WHILE{$C\in \mathcal{C}: C$.will\_merge} 
	\FORALL{$C'\in C.\text{neighbors}$}
		\STATE $C'.\text{update\_dissimilarity}(C, C.\text{get\_dissimilarity($C'$)})$
	\ENDFOR
\ENDWHILE  
\end{algorithmic}
\end{algorithm}

\vspace{-3mm}

\begin{algorithm}[H]
\caption*{\centering \textbf{Update Nearest Neighbors}}
\label{alg:update_nns}
\begin{algorithmic}[1]
\STATE \textbf{Require: }Set of clusters, $\mathcal{C}$
\WHILE{$C\in \mathcal{C}$} 
	\IF{$C$.will\_merge or $C$.nn.will\_merge}
    	\STATE $C$.update\_nn()
	\ENDIF
\ENDWHILE  
\end{algorithmic}
\end{algorithm}

\section{Experimental evaluation}
\label{sec:experiments}

In this section we experimentally evaluate RAC's scalability. In particular, since RAC produces the same hierarchy as HAC, we do not compare RAC's output to any other clustering algorithms.
We run our experiments on a modern cloud infrastructure with access to hundreds of multi-core machines  connected by a fast network. In all experiments below, we share the run-times in a relative scale or as  speedup compared to a baseline. We first demonstrate the validity of our assumptions by investigating the merge characteristics on real datasets. Next we show that RAC can scale to very large real datasets (eg. 1B nodes and 1T edges). We demonstrate that RAC effectively uses the
parallelism available to it by running experiments with different machine/CPU configurations. Finally, we analyze the runtime of the merging phase of RAC, giving insights into its performance characteristic. 
 
\begin{table}[h!]
\small
\caption{Datasets}
\label{tbl:largest_datasets_description}
   \centering
   \begin{tabular}{l|l|l|l}
   \textbf{Dataset} & \textbf{Nodes} & \textbf{Edges} & \textbf{Distance} \\ 
   \hline
   SIFT1B &  1B  & 720B & $l_2$ \\
   SIFT1M & 1M & 1T & $l_2$ \\
   SIFT200K &  200K & 40B & $l_2$ \\
   WEB88M &  88M & 392B & Cosine \\
   News20 &  18846 & 355M & Cosine \\
   RCV1 (training) & 23149 & 0.5B & Cosine \\ 
   \end{tabular}
\end{table}

We study datasets of objects represented as vectors and construct graphs on them using a suitable distance metric. Our datasets are listed in Table \ref{tbl:largest_datasets_description}.
News20 \citep{Lang95} and RCV1 \citep{amini2009learning} are well known medium sized datasets. While they are too small for RAC's benefits to be visible, we use them to study the possibilities of parallelism in RAC. We use the large SIFT datasets \citep{jegou2011searching} and WEB88M to demonstrate the scaling capacities of RAC. Note that the last SIFT dataset (SIFT200K) was constructed for this work by sampling randomly from the SIFT1M dataset. We generated WEB88M by crawling popular web pages and generating bag of words sparse features.

To demonstrate that RAC indeed enables agglomerative clustering at unprecedented scales, we cluster datasets whose sizes are toward the limits of available infrastructure. We experiment with both a) complete graphs on the datasets b) sparse graphs (i.e. graphs with (much) fewer than $n^2$ edges). These sparse graphs are constructed by keeping first $k$ nearest neighbors or only neighbors within an $\epsilon$ ball. In particular, SIFT1M is complete while SIFT1B is sparse. The sparse graph setting is common in practical applications.

\begin{figure}[h!]
\includegraphics[width=1\linewidth]{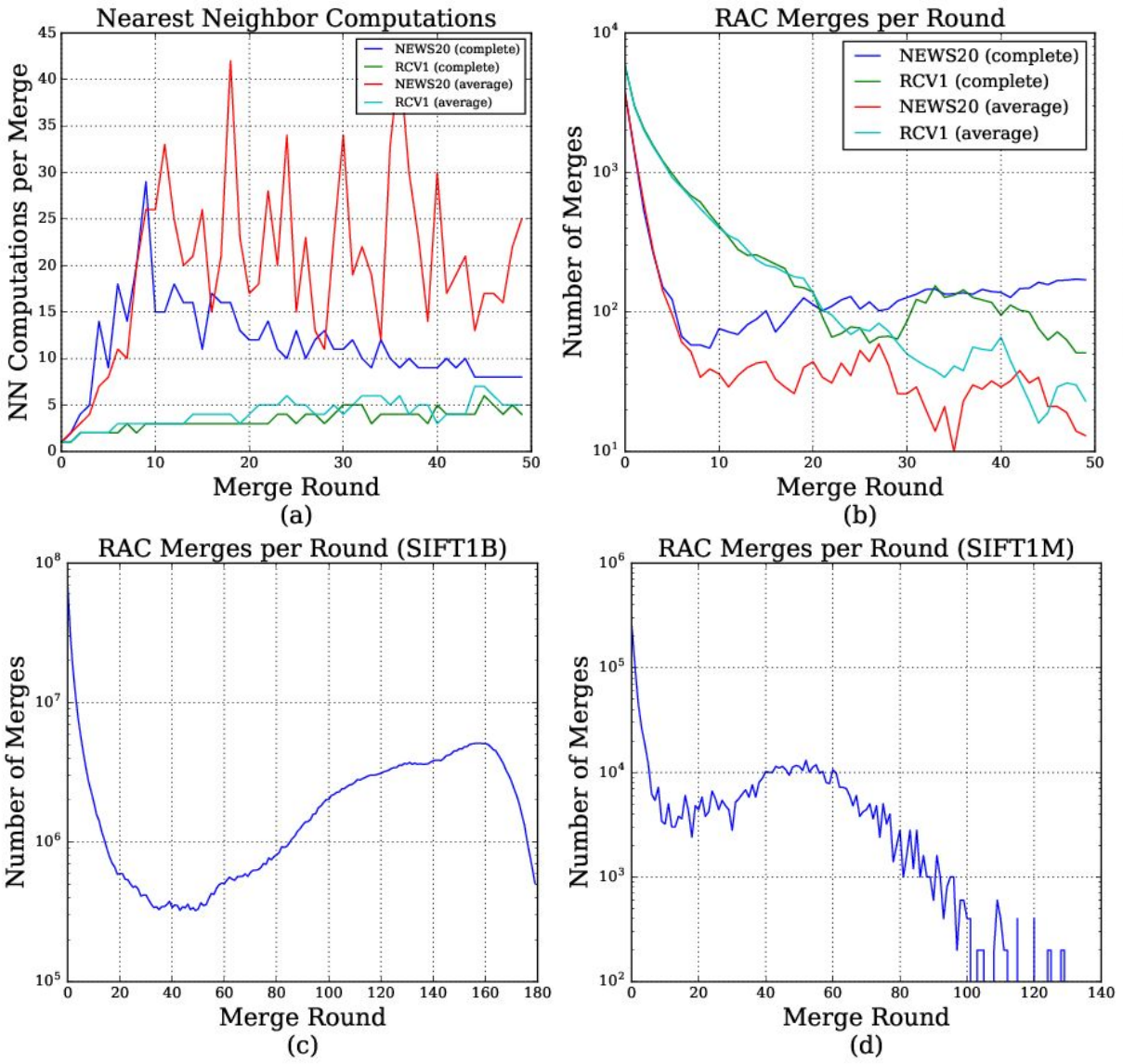}
\caption{Merge characteristics: a) Number of nearest neighbor updates per merge for News20 and RCV1
b) Number of merges per round for News20 and RCV1.  c) and d) Number of merges per round for SIFT1B and SIFT1M respectively. }
 \label{fig:merge_count_for_datasets}
\end{figure}

\textbf{Merge Characteristics ($\alpha$ and $\beta$)}: Recall that RAC proceeds in rounds, merging a number of reciprocal nearest neighbors simultaneously.
Figure \ref{fig:merge_count_for_datasets} shows the number of merges per round for $4$ datasets, demonstrating high parallelism (specially for initial rounds). 
What perhaps is non-intuitive is hump for the SIFT datasets suggesting that RAC goes through a bottleneck before finding opportunities for parallelism again. For News20 and RCV1 
dataset, we also show (Figure \ref{fig:merge_count_for_datasets} (a)) that the number
of nearest neighbor updates per merge is bounded. Recall that this was an important part of our complexity analysis in Theorem \ref{thm:nearly_linear_time} (formulated as parameter $\beta$).

\vspace{-5pt}
\textbf{Experiments on Large Datasets:} Table \ref{tbl:largest_datasets} shows representative timings for the 4 large datasets we achieved for complete linkage.
\begin{table}[h!]
\caption{Performance of RAC on large datasets. }
\label{tbl:largest_datasets}
\centering
  {
      \begin{tabular}{l|l|l|l|l}
      \textbf{Metric} & \textbf{WEB88M} & \textbf{SIFT1B}  &  \textbf{SIFT1M} & \textbf{SIFT200K} \\ \hline
      \# of Machines & 80 & 200 & 200 & 120   \\
      CPUs/Machine  & 16 & 16 & 8 & 4  \\
      Merges  & 69M & 490M & 1M & 200K \\
      Merge Rounds & 170 & 182 &  124 & 112 \\
      Merge Time (relative) & 1.0 & 2.0 & 32.0 & 9 \\
      \end{tabular}
  }
\end{table}
We have specified run times normalized relative to WEB88M's runtime, but note that all experiments finished in under a few hours. The times presented are for merging only and do not include time to load the edges from disk. Edge loading accounts for 15\% to 50\% of the total run time in these datasets.

\begin{figure}[h!]
  \includegraphics[width=0.9\linewidth]{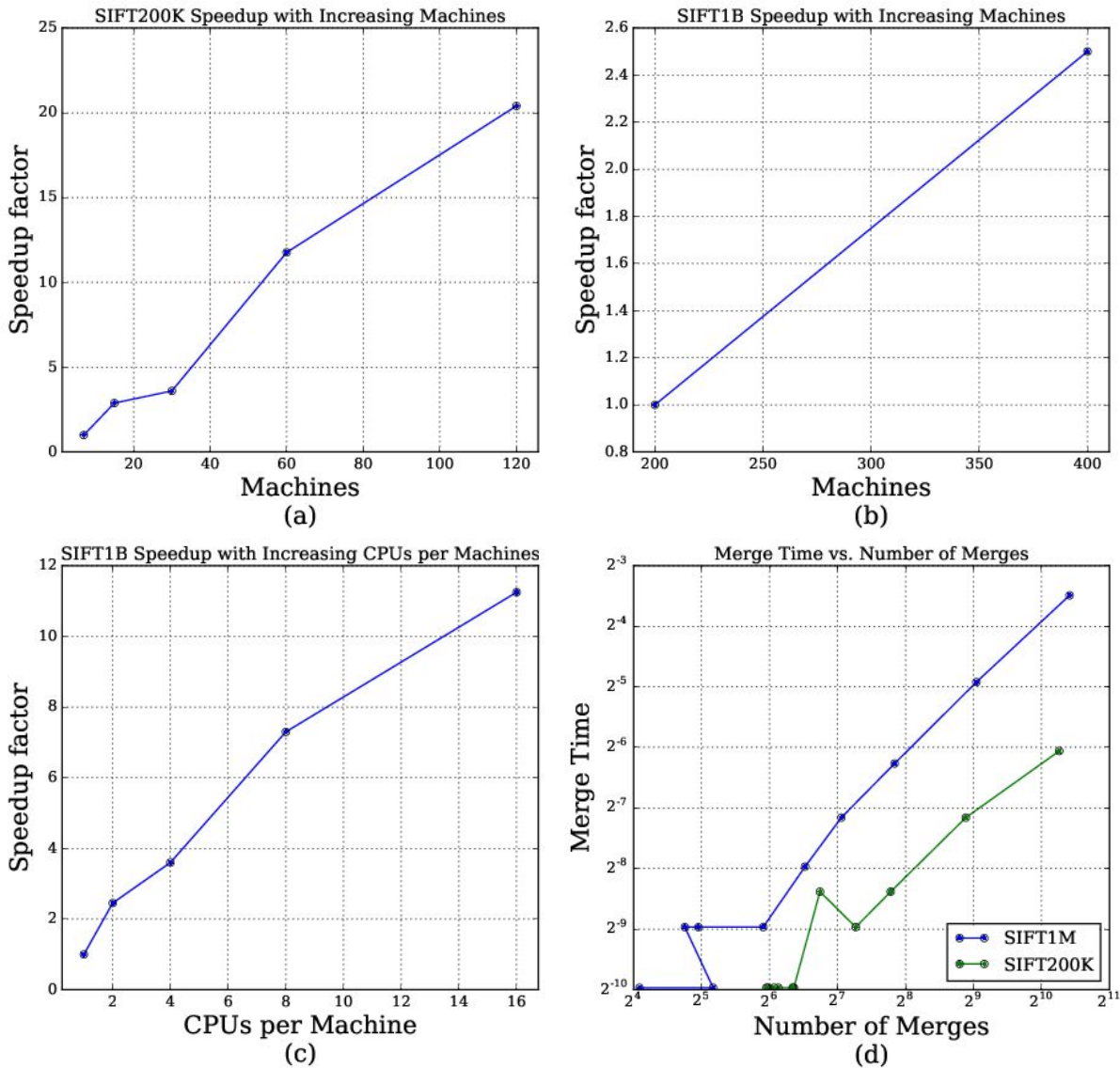}
  \caption{Improved runtimes with increased machines and CPU in Figures \ref{fig:speedup_and_timings}(a-c). Figures \ref{fig:speedup_and_timings}(a) and \ref{fig:speedup_and_timings}(b) show how runtime improves with number of machines for SIFT200K and SIFT1B datasets respectively. Figures \ref{fig:speedup_and_timings}(c) shows speedups on SIFT1B when running RAC using 200 machines but increasing the number CPUs per machine. Figure \ref{fig:speedup_and_timings}(d) shows a log-log plot of merge time as a function of the number of mergers occurring in a round (for SIFT1M and SIFT200K). Merge time scales roughly linearly in number of merges.}
  \label{fig:speedup_and_timings}
\end{figure}

A few observations about the results in Table \ref{tbl:largest_datasets}. First, the number of merge rounds for all datasets are small compared to the size of the dataset, providing further support of the efficacy of the RAC approach. We note that the complete graphs register slower timing (compare SIFT1M vs SIFT1B), which is not surprising since the neighborhoods shuttling across the network is much larger. In Figure \ref{fig:speedup_and_timings} we demonstrate how RAC scales with access to more machines and CPUs, showing that the RAC algorithm is able to effectively use parallelism in the algorithm and the infrastructure to solve HAC at scale.

\vspace{-5pt}
Finally, let us dig a bit deeper into the merge phase of the algorithm.
We already mentioned that neither compute nor network dominates the run-time. This indicates that the merge phase, which uses network and compute equally (see Table \ref{tbl:runtime_breakdown}),  would dominate the runtime. This is indeed what we find experimentally, i.e., that run-time for merge rounds is nearly linear in the number of merges (Figure \ref{fig:speedup_and_timings} (d)) in that round.

\vspace{-5pt}
\section{Conclusion}
We have introduced Reciprocal Agglomerative Clustering (RAC), a distributed variation of Hierarchical Agglomerative Clustering (HAC) that can be efficiently parallelized. RAC achieves efficiency by identifying reciprocal nearest neighbors and merging them in parallel. We have proved the correctness of RAC and that it achieves speedup guarantees in stable cluster trees as well as in certain probabilistic settings. Lastly, we have experimentally shown that RAC can scale to billions of nodes and trillions of edges.

\clearpage
\bibliographystyle{plainnat}
\bibliography{hac}

\begin{thebibliography}{19}
\providecommand{\natexlab}[1]{#1}
\providecommand{\url}[1]{\texttt{#1}}
\expandafter\ifx\csname urlstyle\endcsname\relax
  \providecommand{\doi}[1]{doi: #1}\else
  \providecommand{\doi}{doi: \begingroup \urlstyle{rm}\Url}\fi

\bibitem[Amini et~al.(2009)Amini, Usunier, and Goutte]{amini2009learning}
Massih Amini, Nicolas Usunier, and Cyril Goutte.
\newblock Learning from multiple partially observed views-an application to
  multilingual text categorization.
\newblock In \emph{Advances in neural information processing systems}, pages
  28--36, 2009.

\bibitem[Awasthi et~al.(2014)Awasthi, Balcan, and Voevodski]{awasthi2014local}
Pranjal Awasthi, Maria Balcan, and Konstantin Voevodski.
\newblock Local algorithms for interactive clustering.
\newblock In \emph{International Conference on Machine Learning}, pages
  550--558, 2014.

\bibitem[Bateni et~al.(2017)Bateni, Behnezhad, Derakhshan, Hajiaghayi, Kiveris,
  Lattanzi, and Mirrokni]{bateni2017affinity}
MohammadHossein Bateni, Soheil Behnezhad, Mahsa Derakhshan, MohammadTaghi
  Hajiaghayi, Raimondas Kiveris, Silvio Lattanzi, and Vahab Mirrokni.
\newblock Affinity clustering: Hierarchical clustering at scale.
\newblock In \emph{Advances in Neural Information Processing Systems}, pages
  6867--6877, 2017.

\bibitem[Cathey et~al.(2007)Cathey, Jensen, Beitzel, Frieder, and
  Grossman]{cathey2007exploiting}
Rebecca~J Cathey, Eric~C Jensen, Steven~M Beitzel, Ophir Frieder, and David
  Grossman.
\newblock Exploiting parallelism to support scalable hierarchical clustering.
\newblock \emph{Journal of the Association for Information Science and
  Technology}, 58\penalty0 (8):\penalty0 1207--1221, 2007.

\bibitem[De~Rham(1980)]{de1980classification}
C~De~Rham.
\newblock La classification hi{\'e}rarchique ascendante selon la m{\'e}thode
  des voisins r{\'e}ciproques.
\newblock \emph{Les Cahiers de l’Analyse des Donn{\'e}es}, 135:\penalty0 144,
  1980.

\bibitem[Defays(1977)]{defays1977efficient}
Daniel Defays.
\newblock An efficient algorithm for a complete link method.
\newblock \emph{The Computer Journal}, 20\penalty0 (4):\penalty0 364--366,
  1977.

\bibitem[Doob(1953)]{doob1953stochastic}
J.L. Doob.
\newblock \emph{Stochastic Processes}.
\newblock Wiley Publications in Statistics. John Wiley \& Sons, 1953.

\bibitem[J{\'e}gou et~al.(2011)J{\'e}gou, Tavenard, Douze, and
  Amsaleg]{jegou2011searching}
Herv{\'e} J{\'e}gou, Romain Tavenard, Matthijs Douze, and Laurent Amsaleg.
\newblock Searching in one billion vectors: re-rank with source coding.
\newblock In \emph{Acoustics, Speech and Signal Processing (ICASSP), 2011 IEEE
  International Conference on}, pages 861--864. IEEE, 2011.

\bibitem[Jeon and Yoon(2015)]{jeon2015multi}
Yongkweon Jeon and Sungroh Yoon.
\newblock Multi-threaded hierarchical clustering by parallel nearest-neighbor
  chaining.
\newblock \emph{IEEE Transactions on Parallel and Distributed Systems},
  26\penalty0 (9):\penalty0 2534--2548, 2015.

\bibitem[Lance and Williams(1967)]{lance1967general}
Godfrey~N Lance and William~Thomas Williams.
\newblock A general theory of classificatory sorting strategies: 1.
  hierarchical systems.
\newblock \emph{The computer journal}, 9\penalty0 (4):\penalty0 373--380, 1967.

\bibitem[Lang(1995)]{Lang95}
Ken Lang.
\newblock Newsweeder: Learning to filter netnews.
\newblock In \emph{Proceedings of the Twelfth International Conference on
  Machine Learning}, pages 331--339, 1995.

\bibitem[Moseley and Wang(2017)]{moseley2017approximation}
Benjamin Moseley and Joshua Wang.
\newblock Approximation bounds for hierarchical clustering: Average linkage,
  bisecting k-means, and local search.
\newblock In \emph{Advances in Neural Information Processing Systems}, pages
  3097--3106, 2017.

\bibitem[Murtagh(1983)]{murtagh1983survey}
Fionn Murtagh.
\newblock A survey of recent advances in hierarchical clustering algorithms.
\newblock \emph{The Computer Journal}, 26\penalty0 (4):\penalty0 354--359,
  1983.

\bibitem[Murtagh(1984)]{murtagh1984complexities}
Fionn Murtagh.
\newblock Complexities of hierarchic clustering algorithms: state of the art.
\newblock \emph{Computational Statistics Quarterly}, 1\penalty0 (2):\penalty0
  101--113, 1984.

\bibitem[Rammal et~al.(1985)Rammal, d'Auriac, and
  Dou{\c{c}}ot]{rammal1985degree}
R~Rammal, JC~Angles d'Auriac, and Benoˆ{\i}t Dou{\c{c}}ot.
\newblock On the degree of ultrametricity.
\newblock \emph{Journal de Physique Lettres}, 46\penalty0 (20):\penalty0
  945--952, 1985.

\bibitem[Rasmussen and Willett(1989)]{rasmussen1989efficiency}
Edie~M Rasmussen and Peter Willett.
\newblock Efficiency of hierarchic agglomerative clustering using the icl
  distributed array processor.
\newblock \emph{Journal of Documentation}, 45\penalty0 (1):\penalty0 1--24,
  1989.

\bibitem[Sibson(1973)]{sibson1973slink}
Robin Sibson.
\newblock Slink: an optimally efficient algorithm for the single-link cluster
  method.
\newblock \emph{The computer journal}, 16\penalty0 (1):\penalty0 30--34, 1973.

\bibitem[Sokal(1958)]{sokal1958statistical}
Robert~R Sokal.
\newblock A statistical method for evaluating systematic relationship.
\newblock \emph{University of Kansas science bulletin}, 28:\penalty0
  1409--1438, 1958.

\bibitem[Williams and Lambert(1966)]{williams1966multivariate}
WT~Williams and JM~t Lambert.
\newblock Multivariate methods in plant ecology: V. similarity analyses and
  information-analysis.
\newblock \emph{The Journal of Ecology}, pages 427--445, 1966.

\end{thebibliography}

\clearpage

{\LARGE
\begin{center}
\textbf{Supplementary material}
\end{center}
}

\section*{}

\vspace{-1.0cm}
We provide proofs of Theorems 4 and 8 in the following sections.
\section*{Proof of Theorem 4}
\begin{proof}
We fix $n\in \mathbb{N}$ and use the notation $\epsilon=2^{-4n}$, $P_k = (k+1) + \epsilon (k+1)^2$, and $X = \{P_i|i=0, \ldots, 2^n-1\}$.

We first prove that RAC with average linkage takes $\Omega(2^n)$ rounds with input $X$. We do this by showing that in each round, at most one merge involves the singleton clusters $\{P_i\}$ (in fact, we show that they are merged in order of increasing $i$). We first note that clusters always consist of a contiguous set of points. In a given round, let $r$ be the smallest index such that for $i\geq r$, each $P_i$ is in a singleton cluster. Let $C_r=\{\{P_i\}| i = r, \ldots, 2^n-1\}$ denote these singleton clusters. For $i > r$, cluster $\{P_i\}$'s nearest neighbor is $\{P_{i-1}\}$. As for $\{P_r\}$, because $P_{r-1}$ is not in a singleton cluster, $\{P_r\}$'s nearest neighbor must be $\{P_{r+1}\}$. As a result, the only reciprocal nearest neighbor involving clusters in $C_r$ is the pair $(\{P_r\}, \{P_{r+1}\})$. This completes the proof of the lower bound.

In the rest of the proof, we show that HAC (and hence RAC) with average linkage produces a dendrogram of height $n$ with $X$ as input. Consider the natural complete binary tree $\mathcal{T}$ on $X$. Specifically, 
define nodes at the bottom level $N^0_i = P_i$.
For higher levels, define $N^l_i$ recursively, i.e. $N^l_i$ will have $N^{l-1}_{2i}$ and $N^{l-1}_{2i + 1}$ as its children. 

We note following simple properties of $\mathcal{T}$ below, where we use the notation $X > Y$ for sets $X$ and $Y$ to mean that every point of $X$ is greater than every point in $Y$. 

\begin{enumerate}
\item $\mathcal{T}$ has height $n$.
\item There are $2^{n-l}$ nodes at level $l$.
\item  The subtree rooted at $N^l_i$ contains the following $2^l$ points as leaves:\\
$\{P_k | i \cdot 2^l \leq k \leq (i + 1) \cdot 2^l - 1 \}$. 
\item $N^l_{i + 1} > N^l_i$.
\end{enumerate}

We will show that HAC with average linkage produces $\mathcal{T}$. Since $\mathcal{T}$ has height $n$, this will complete the proof.

Let $d(P_k, P_i) =|P_k - P_i|$ be the Euclidean distance between two points. For sets $X$ and $Y$, define their distance according to average linkage, i.e., $d(X, Y) = \frac{1}{|X||Y|} \sum_{P_k \in X, P_i \in Y} d(P_k, P_i)$. The following is easy to see:
\begin{lemma}
Let $X, Y, Z$ be such that $X > Y > Z$. Then $d(X, Y) < d(X, Z)$.
\end{lemma}

We will need the following series of somewhat tedious but straightforward lemmas regarding the distances between pairs of nodes in $\mathcal{T}$.
\begin{lemma}
$d(N^l_i, N^l_{i + 1})$ is monotonically increasing in $i$.
\label{perf:lemma:monotone}
\end{lemma}
\begin{proof}
We evaluate the more general distance between nodes $N^l_i$ and $N^l_{i + u}$ in the same level $l$ for any $u > 0$:

\begin{align}
& d(N^l_i, N^l_{i+u}) = \frac{1}{2^{2l}}\sum^{2^l-1}_{r = 0} \sum^{2^l-1}_{s = 0} d\left(P_{2^l\cdot i + r}, P_{2^l\cdot (i + u) + s}\right) \nonumber \\
&= \frac{1}{2^{2l}}\sum_{r, s=0}^{2^l-1} (2^l (i+u) \cdot s + (2^l (i+u) \cdot s)^2 \epsilon -  2^l i+ \cdot r - (2^l i+ \cdot r)^2 \epsilon)  \nonumber  \\
&= \frac{1}{2^{2l}} \sum_{r,s=0}^{2^l-1} ( (2^l \cdot u + s - r) + (2^l \cdot u + s - r)\cdot(2^{l + 1} i + 2^l \cdot u +  s - r) \cdot \epsilon)
\label{eqn:adjacent_nodes}
\end{align}

Setting $u = 1$ and noting that $s$ and $r$ are upper bounded by $2^l - 1$, we have that $(2^l + s - r)$ is positive, and hence the last expression is monotonically increasing in $i$.
\end{proof}

The following lemma relates the nodes in two different levels:

\begin{lemma}
Consider the nodes $N^l_{2 i+2}$ and $N^{l+1}_j$ for $j \leq i$. 
Then $d(N^l_{2 i+2}, N^l_{2 i+3}) < d(N^l_{2 i+2}, N^{l+1}_j)$. Informally, the distance from a node to its neighbor at the same level on its right is less than the distance to its neighbor on a higher level on its left.
\label{perf:lemma:same_level_better}
\end{lemma}
\begin{proof}
Since $N^{l+1}_{j-1} <  N^{l+1}_j$, it suffices to prove the lemma for $j = i$. Now, by construction,
$N^{l+1}_i$ is the result of the merging $N^{l}_{2 i}$ and $N^{l}_{2 i + 1}$ and thus

\begin{equation*}
d(N^{l+1}_i, N^l_{2 i+2}) = \frac{1}{2} (d(N^{l}_{2 i}, N^l_{2 i+2}) + d( N^{l}_{2 i + 1}, N^l_{2 i+2})).
\end{equation*}
Thus, we need to prove that $\frac{1}{2} (d(N^{l}_{2 i}, N^l_{2 i+2}) + d( N^{l}_{2 i + 1}, N^l_{2 i+2})) > d(N^l_{2 i+2}, N^l_{2 i+3})$, which (replacing $2i$ with $i$) is equivalent to proving
\begin{equation*}
\frac{1}{2} (d(N^{l}_{i}, N^l_{i+2}) + d( N^{l}_{i + 1}, N^l_{i+2})) > d(N^l_{i+2}, N^l_{i+3})    
\end{equation*}
for any $i$ and $l$. Rewriting each distance term using Equation (\ref{eqn:adjacent_nodes}), it suffices to prove for any $r$ and $s$ that

\begin{align*}
 & (2^l \cdot 2 + s - r) + (2^l \cdot 2 + s - r)\cdot(2^{l + 1} i + 2^l \cdot 2 +  s - r) \cdot \epsilon \nonumber \\
 & +  (2^l + s - r) + (2^l + s - r)\cdot(2^{l + 1} (i + 1) + 2^l  +  s - r) \cdot \epsilon > \nonumber \\
  & 2  ((2^l  + s - r) + (2^l u + s - r)\cdot(2^{l + 1} (i + 2 )+ 2^l  +  s - r) \cdot \epsilon).
 \end{align*}

Focusing first on the terms not involving $\epsilon$, we have
\begin{align*}
(2^l \cdot 2 + s - r) + (2^l + s - r) - 2  (2^l  + s - r) &= (2^l \cdot 2 + s - r) -  (2^l  + s - r) \\
&= 2^l \geq 1.
\end{align*}
Since terms including $\epsilon$ are much smaller than $1$, the lemma follows.
\end{proof}

Lastly, this lemma states that adjacent distances at a higher level are
larger than adjacent distances at a lower level. 
\begin{lemma}
If $l > l'$, then $d(N^l_i, N^l_{i+1}) > d(N^{l'}_j, N^{l'}_{j+1})$ 
\label{perf:lemma:diff_levels}
\end{lemma}
\begin{proof}
Using Equation (\ref{eqn:adjacent_nodes}) with $u = 1$, we have 
\begin{align*}
d(N^l_i, N^l_{i+1}) &= \frac{1}{2^{2l}} \sum_{r,s=0}^{2^l-1} (2^l + s - r) + O(\epsilon) \\
&=2^l + \frac{1}{2^{2l}} \sum_{r,s=0}^{2^l-1} (s - r) + O(\epsilon).
\end{align*}
Note that the second term vanishes by symmetry. Thus, $d(N^l_i, N^l_{i+1}) = 2^l + O(\epsilon)$ and for $l'\geq l+1$,  $d(N^{l'}_j, N^{l'}_{j+1}) - d(N^l_i, N^l_{i+1}) = 2^{l'} - 2^l + O(\epsilon) \geq 2^l + O(\epsilon) > 0$ for sufficiently small epsilon. To see that our choice of $\epsilon = 2^{-4n}$ suffices, we can bound the term of Equation (\ref{eqn:adjacent_nodes}) involving $\epsilon$ as
\begin{equation*}
(2^l+ s - r)\cdot(2^{l + 1} i + 2^l +  s - r) \cdot \epsilon < 2^{4n-1} \epsilon, 
\end{equation*}
where we have used $i\leq 2^{n-l}$, $s-r \leq 2^l - 1$, and $l\leq n-1$. Since the average of these terms is also bounded by the same quantity, the proof is complete.
\end{proof}

We now show that HAC with average linkage produces $\mathcal{T}$. Consider the state $S$ of the clustering algorithm after a merge. It consists of a partition of the nodes into a number of
sets. We define $S$ to be \textbf{good} if there exists $l$ and $k$ with $0 \leq k < 2^{n-l}$ such that
\begin{align}
S = \{N^l_i : i < k\} \cup \{N^{l-1}_j : j \geq 2k\}.
\end{align}
Informally, this is a clustering of nodes covered by one or two levels where the nodes in each level consists of a single contiguous span. Note that the case $k = 0$ corresponds to a single layer.

We claim that after each step of merging in HAC, the state is \textbf{good}. Note that this shows that $\mathcal{T}$ is indeed the tree produced by HAC with average linkage.

We proceed by induction. The base case is clear, since the state consisting of each $P_i$ as a single node is \textbf{good}.
Now, consider an arbitrary state for a given $l$ and $k$. We claim that nodes $N^{l-1}_{2k}$ and $N^{l-1}_{2k + 1}$ will be merged next. The fact that no other pair at level $l-1$ will be merged is implied by Lemma \ref{perf:lemma:monotone}. Lemma \ref{perf:lemma:diff_levels} implies that no pair at level $l$ is merged. Finally, a pair in different levels (one in $l$ and another in $l-1$) merging is ruled out by Lemma \ref{perf:lemma:same_level_better}. It is clear that the merge of
$N^{l-1}_{2k}$ and $N^{l-1}_{2k + 1}$ results in a \textbf{good} state, completing the proof.

\end{proof}

\section*{Proof of Theorem 8}
We prove that for the bounded probabilistic graph model, the expected runtime of RAC is $O(n^2)$.
\begin{proof}
Since the total number of rounds is $O(\log n)$ in expectation, and each round takes $O(n^2)$ time, we easily obtain the weaker expected runtime guarantee of $O(n^2\log n)$. The following analysis strengthens this result by shaving off the $\log n$ factor. Let $n_j$ be the number of clusters at the end of round $j$, with $n_0 = n$, and let $X_j = n_j/n_{j-1}$ be the fraction by which the number of clusters reduces in round $j$. Thus, $n_j = n\prod_{i=1}^j X_i$. Since the number of merges, $M_j$, in each round satisfies $E[M_j] \geq \alpha n_{j-1}$, and $n_j = n_{j-1} - M_j$, it follows that $E[X_j] \leq 1 - \alpha$. A key component of the proof is bounding the variance of the $X_j$'s:
\begin{lemma}
\label{lem:var_bound}
$\Var(X_j|X_1, \ldots, X_{j-1}) \leq B/n_{j-1}$, where $B$ is a constant.
\end{lemma}

Assuming Lemma \ref{lem:var_bound}, the rest of the proof is as follows. From the equation $E[X^2] = E[X]^2 + \Var(X)$, we have $E[X_j^2|X_1, \ldots, X_{j-1}] \leq (1-\alpha)^2 + B/n_{j-1}$. Define $k$ such that $n_{k+1} < B/\alpha \leq n_k$. We decompose RAC's runtime into two parts - the time taken for the first $k$ rounds, and the remaining time. Using the fact that the number of points after the first $k$ rounds is the constant $B/\alpha$, we can write the expected runtime of RAC as
\begin{align*}
E[\sum_{j=0}^k O(n_j)^2 + O(1)] &= \sum_{j=0}^k E[O(n\prod_{i=1}^j X_i)^2] \\
&= O(n^2)\sum_{j=0}^k E[\prod_{i=1}^j X_i^2],
\end{align*}
where we use the convention that the empty product is $1$.
It suffices to show that $\sum_{j=0}^k E[\prod_{i=1}^j X_i^2] = O(1)$. Using Lemma \ref{lem:var_bound}, we have
\begin{align*}
    E[X_j^2|X_1, \ldots, X_{j-1}] &\leq (1-\alpha)^2 + B/n_{j-1} \\
    &\leq (1-\alpha)^2 + \alpha &\text{ (using } B/\alpha \leq n_{j-1}\text{)}\\
    &= 1-\alpha+\alpha^2 < 1.
\end{align*}
Defining $\rho = 1-\alpha+\alpha^2$, we have
\begin{align*}
E[\prod_{i=1}^j X_i^2] &= E[\prod_{i=1}^{j-1} X_i^2E[X_j^2|X_1, \ldots, X_{j-1}]] &\text{ (by law of total expectation)}\\
&\leq \rho E[\prod_{i=1}^{j-1} X_i^2]\\
&\leq \rho^j &\text{ (by induction)}
\end{align*}
and $\sum_{j=0}^k E[\prod_{i=1}^j X_i^2] \leq  \sum_{j=0}^{\infty}\rho^j = 1/(1-\rho) = O(1)$.
\end{proof}
\begin{proof}[Proof of Lemma \ref{lem:var_bound}]
The main idea behind the bound is that the number of merges, $M$, is the sum of Bernoulli random variables with most pairs having non-positive covariance. Let $G = (V, E)$ be a graph with $m$ edges $e_1, \ldots, e_m$ and vertex degrees $d_v \leq d$ for $v\in V$. Note that the bounded degree property implies the edge bound $m \leq d|V|/2$. We assign weights to the edges so that they are sorted at random. For $i = 1, \ldots, m$ let $Y_i$ be the indicator random variable for the event that edge $e_i$ is merged (i.e., the weight of $e_i$ is smaller than the weights of its adjacent edges). Then $M = \sum_{i=1}^m Y_i$ and $\Var(M) = \sum_{i=1}^m \Var(Y_i) + 2\sum_{i<j}\Cov(Y_i, Y_j)$. To bound the pairwise covariances, $\Cov(Y_i, Y_j)$, we consider three cases:
\begin{enumerate}
    \item $e_i$ and $e_j$ are adjacent (i.e., have a vertex in common): In this case, $E[Y_i Y_j] = 0$ since both $e_i$ and $e_j$ cannot be merged. Thus, $\Cov(Y_i, Y_j) = -E[Y_i]E[Y_j] < 0$.
    \item $e_i$ and $e_j$ are not adjacent to a common edge: In this case, $Y_i$ and $Y_j$ are independent, since the random variables that define them are disjoint. Hence, $\Cov(Y_i, Y_j) = 0$.
    \item $e_i$ and $e_j$ are adjacent to a common edge: In this case, we can bound the number of such pairs by $d^2m$ since each common edge $e_i = (a, b)$ gives rise to at most $d_a d_b \leq d^2$ such pairs for $i = 1, \ldots, m$.
\end{enumerate}
Breaking up the covariance term $2\sum_{i < j}\Cov(Y_i, Y_j)$ according to the above cases, we have
\begin{align*}
    \Var(M) &= \sum_{i=1}^m \Var(Y_i) + 2\sum_{i < j}\Cov(Y_i, Y_j)\\
    &\leq m + d^2m = (d^2+1)m\\
    &\leq (d^2+1)dn_{j-1}/2,
\end{align*}
where we have used the fact that for the indicator random variables $Y_i$, $\Var(Y_i)\leq 1$ and $\Cov(Y_i, Y_j)\leq 1$, and that $m \leq dn_{j-1}/2$.

We apply this in the case $M=M_j$. Since $X_j = 1 - M_j/n_{j-1}$, we have 
\begin{align*}
\Var(X_j|X_1, \ldots, X_{j-1}) &= \Var(M_j/n_{j-1})\\
&\leq (1/n_{j-1}^2)\Var(M_j)\\
&\leq (d^2+1)d/(2n_{j-1}).
\end{align*}
\end{proof}

\end{document}